\documentclass[lettersize, 10pt, journal, twoside]{IEEEtran}  

\usepackage{lettrine}
\usepackage{graphicx} 
\usepackage{bm}
\usepackage{multirow}
\usepackage{tabularx}
\usepackage{booktabs}
\usepackage{subcaption}  
\usepackage{mathtools}
\usepackage{tikz}
\usepackage{amsmath}
\usepackage{amssymb}
\usepackage[colorlinks,
linkcolor=red,
anchorcolor=blue,
citecolor=green,
]{hyperref}

\hypersetup{
	colorlinks=true,
	linkcolor=red,
	filecolor=blue,      
	urlcolor=blue
}
\newcommand{\urlnote}[1]{%
  \footnote{\urlstyle{same}\url{#1}}%
}
\usepackage[subtle]{savetrees}
\usepackage{makecell}
\usepackage{pifont}
\usepackage{xcolor}
\definecolor{olivegreen}{HTML}{3C8031}
\definecolor{tableblue}{HTML}{2D2F92}
\definecolor{mygreen}{RGB}{69,144,26}  
\definecolor{mygreenborder}{RGB}{46,96,20}    
\definecolor{myred}{RGB}{146,0,28} 
\definecolor{buttonred}{RGB}{200,28,50}
\definecolor{buttongreen}{RGB}{88,141,49}
\newcommand{\xmark}{\ding{55}}%
\newcommand{\redcross}{\textcolor{red}{\xmark}}

\newcommand{\cmark}{\ding{51}}%
\newcommand{\greencheck}{\textcolor{olivegreen}{\cmark}}

\title{\LARGE \bf
CaFe-TeleVision: A Coarse-to-Fine Teleoperation System with Immersive Situated Visualization for Enhanced Ergonomics
}

\author{Zixin Tang, Yiming Chen, Quentin Rouxel, Dianxi Li, Shuang Wu, and Fei Chen, \textit{Senior Member, IEEE}
\thanks{Manuscript received 21 July 2025; revised 17 October 2025; accepted 14 November 2025. This paper was recommended for publication by Editor Ki-Uk Kyung upon evaluation of the Associate Editor and Reviewers' comments. 
This work was supported in part by the Research Grants Council of the Hong Kong SAR under Grant 14211723, 14222722, 24209021 and C7100-22GF, in part by CUHK \& HUAWEI Foundation Models and Interactive Intelligence Innovation Laboratory TH2520452 and in part by InnoHK of the Government of Hong Kong via the Hong Kong Centre for Logistics Robotics. \textit{(Corresponding authors: Fei Chen)}}
\thanks{Zixin Tang, Yiming Chen,  Quentin Rouxel, Dianxi Li, and Fei Chen are with the Department of Mechanical and Automation Engineering, T-Stone Robotics Institute, The Chinese University of Hong Kong, Hong Kong SAR (email: {zxtang@mae.cuhk.edu.hk, ymchen@mae.cuhk.edu.hk, \mbox{quentinrouxel}@cuhk.edu.hk, dxli@mae.cuhk.edu.hk, f.chen@ieee.org}).}
\thanks{Shuang Wu is with Huawei Hong Kong Research Center (email: {wushuangust@gmail.com}).}
\thanks{Digital Object Identifier (DOI): see top of this page.}
}

\begin{document}

\makeatletter
\let\@oldmaketitle\@maketitle%
\renewcommand{\@maketitle}{\@oldmaketitle%
\centering
\vspace{-0mm}
\includegraphics[width=2\columnwidth]{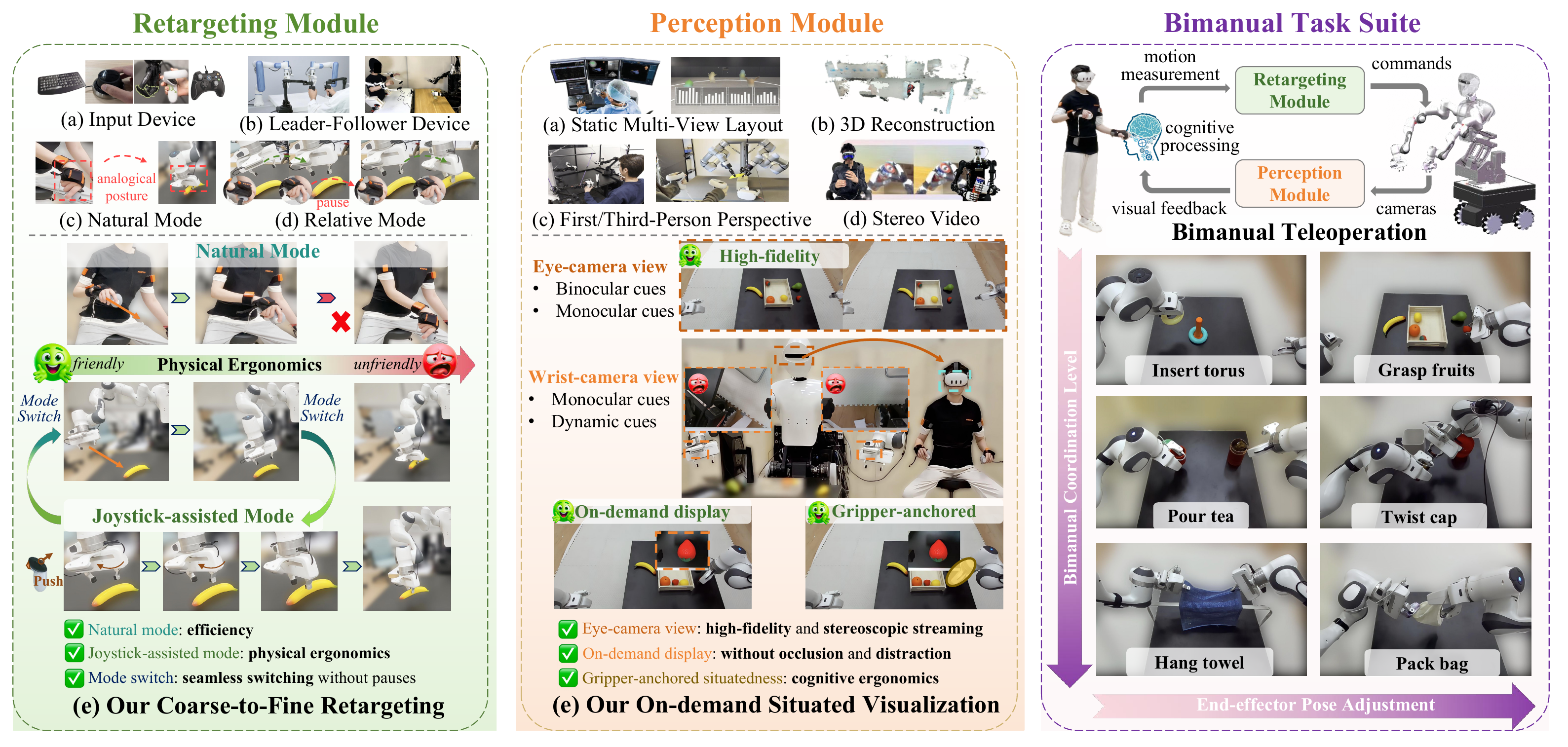}
\setcounter{figure}{0}
    \captionof{figure}{
    \textit{Left and Middle}: Comparison of various teleoperation system designs for retargeting and perception module, respectively. Unlike conventional approaches \textit{(a)-(d)}, \textbf{CaFe-TeleVision} focuses on jointly enhancing efficiency, physical ergonomics, and cognitive ergonomics. \textit{Left (e)} presents our coarse-to-fine retargeting mechanism that preserves physical ergonomics via seamless mode switching. \textit{Middle (e)} shows our novel on-demand situated visualization technique to reduce eye focus shift, visual distraction, and occlusion, thereby improving cognitive ergonomics during multi-view processing. \textit{Right}: Six tasks for evaluating teleoperation systems, spanning different bimanual coordination levels and end-effector pose adjustment requirements.}
\label{fig: overview}
\vspace{-7mm}
}
\makeatother
\maketitle


\begin{abstract}
Teleoperation presents a promising paradigm for remote control and robot proprioceptive data collection. Despite recent progress, current teleoperation systems still suffer from limitations in efficiency and ergonomics, particularly in challenging scenarios. In this paper, we propose CaFe-TeleVision, a coarse-to-fine teleoperation system with immersive situated visualization for enhanced ergonomics. At its core, a coarse-to-fine control mechanism is proposed in the retargeting module to bridge workspace disparities, jointly optimizing efficiency and physical ergonomics. To stream immersive feedback with adequate visual cues for human vision systems, an on-demand situated visualization technique is integrated in the perception module, which reduces the cognitive load for multi-view processing. The system is built on a humanoid collaborative robot and validated with six challenging bimanual manipulation tasks. 
User study among 24 participants confirms that CaFe-TeleVision enhances ergonomics with statistical significance, indicating a lower task load and a higher user acceptance during teleoperation. Quantitative results also validate the superior performance of our system across six tasks, surpassing comparative methods by up to $28.89$\% in success rate and accelerating by $26.81$\% in completion time. Project webpage: \url{https://clover-cuhk.github.io/cafe_television/}
\end{abstract}

\section{Introduction}
\lettrine{T}{eleoperation}, one of the earliest research areas in robotics, offers a promising paradigm to allow human operators to remotely control robots to perform various tasks in different environments, such as medical surgery and space exploration \cite{sheridan1989telerobotics}. With the rise of data-driven skill learning \cite{tang2023csgp, chi2023diffusion, kim2024openvla, rouxel2025extremumflowmatching}, teleoperation systems have gained prominence as an effective demonstration collection approach \cite{schwarz2023robust, zhao2023learning, ding2024bunny, liu2025avr}.

Typically, teleoperation systems include two modules: a retargeting module that maps human motions to robot control references, and a perception module that obtains perceptual feedback from robot side \cite{darvish2023teleoperation}. In retargeting module, task-space mapping \cite{ding2024bunny, rouxel2022multicontact} converts human hand poses into target poses for robot end-effectors, allowing more flexibility than joint-space mapping \cite{zhao2023learning}. In perception module, many studies rely on 2D monitors \cite{mandlekar2018roboturk} or first/third-person perspectives \cite{ozdamar2022shared, lin2024learning} for feedback, which either lack stereoscopic perception \cite{fung2005case, livatino2009stereo} or hinder remote teleoperation. To mitigate these limitations, streaming stereo video to a Virtual Reality (VR) head-mounted display (HMD) provides an effective solution \cite{ishiguro2018high, penco2019multimode, cheng2024open}.

Despite recent progress, two challenges remain inadequately addressed in existing approaches, which compromise teleoperation performance in efficiency and ergonomics. 
First, morphological and kinematic discrepancies between humans and robots introduce the workspace mismatch problem in task-space mapping. This issue is particularly prominent in orientation workspace, due to the limited rotational range of the operator’s anatomical joints (e.g., ulnar/radial deviation) \cite{ryu1991functional}. Improper retargeting induces muscle strain in teleoperation, adverse to physical ergonomics \cite{mehta2016integrating}.
Second, stereo video streams from robot eye cameras fail to provide adequate visual cues for human vision systems, such as dynamic cues (motion parallax) \cite{el2019survey}. This limitation is troubled by occlusion-induced blind spots, leading to poor cognitive ergonomics \cite{mehta2016integrating} and ultimately impairing vision-based decision-making.

We aim to bridge the challenges in a practical and easily deployable manner. To this end, we propose CaFe-TeleVision, a coarse-to-fine teleoperation system with immersive situated visualization for enhanced ergonomics, as shown in Fig~\ref{fig: overview}. At its core, a coarse-to-fine control mechanism is proposed to bridge workspace disparities through two complementary modes: natural mode and joystick-assisted mode.
Specifically, the natural mode follows the pattern of scaling position and aligning orientation to retarget human wrist motions into analogous target poses for end-effectors. This intuitive analogy allows operators to perceive robots as kinematic avatars for efficient operation. 
When encountering kinematic limits due to orientation mismatch, they can seamlessly switch to the joystick-assisted mode for incremental refinement. The dual modes jointly improve efficiency and physical ergonomics.

In addition, stereoscopic visual feedback from the eye camera is streamed in real time as the primary display, complemented by two wrist cameras. Unlike static multi-view layout, an immersive situated visualization technique is integrated in perception module. Specifically, situated visualization is an emerging concept within visualization, which means displaying data in spatial proximity with its contextual referent (i.e., situatedness) \cite{bressa2021s}. To reduce eye focus shift, visual distraction, and occlusion, our system implements on-demand display and gripper-anchored situatedness. The operator can control the visibility of wrist views through signals of VR controllers and visualize them spatially coupled to the gripper. These designs enhance cognitive ergonomics during multi-view processing. 

The system is evaluated on a humanoid collaborative robot with six bimanual tasks, covering various collaborative levels and operational difficulty (see Fig.~\ref{fig: overview}). Experiments in the user study confirm that CaFe-TeleVision significantly improves ergonomics, with a lower task load and a higher user acceptance during teleoperation. Quantitative results also validate its superior performance, boosting success rates by up to $28.89$\% and efficiency by $26.81$\%.

\section{Related Works}

\subsection{Measurement and Feedback in Teleoperation}

Teleoperation systems utilize various approaches to measure human motion information and transmit perceptual feedback \cite{sheridan1989telerobotics}. Table~\ref{tab:interfaces} summarizes their characteristics, focusing mainly on kinematic measurement and visual feedback. Fig.~\ref{fig: overview} presents examples. A more comprehensive review can be found in \cite{darvish2023teleoperation}. Typically, motion measurement methods use user interaction or measurement algorithms for estimation, which can be classified into five categories: input device (e.g., keyboard, mouse), Leader-Follower device (e.g., exoskeleton, Da Vinci), vision-based detector (e.g., OpenPose \cite{martinez2019openpose}), optical tracking (e.g., OptiTrack), and IMU sensor (e.g., Xsens \cite{roetenberg2009xsens}). In comparison, IMU sensors show relatively higher performance with respect to intuitiveness, portability, occlusion robustness, and precision. For visual feedback, operators always observe the workspace through GUI monitors, first/third-person perspective, or VR HMDs (e.g., Meta Quest). Recently, VR HMDs are sprouting as an effective solution, replicating immersive stereo vision via 3D reconstruction \cite{audonnet2024immertwin, patil2024radiance} or stereo streaming \cite{schwarz2023robust, cheng2024open}, where the latter maintains higher fidelity.

\begin{table}[t]
\caption{Characteristics of motion measurement and visual feedback approaches.}
\label{tab:interfaces}
\vspace{-2.5mm}
\begin{center}
\setlength{\tabcolsep}{2pt}
\begin{tabular}{ccccc}
\toprule
\multirow{2}{*}{Motion Measurement} & \multirow{2}{*}{Intuitive} & \multirow{2}{*}{Portable} & Occlusion- & High- \\
& & & robust & precision\\
\midrule
Input Device & \redcross & \greencheck & \greencheck & \redcross  \\
Leader-Follower Device & \greencheck & \redcross & \greencheck & \greencheck \\
Vision-based Detector & \greencheck & \greencheck & \redcross & \redcross \\
Optical Tracking & \greencheck & \redcross & \redcross & \greencheck  \\ 
IMU Sensor & \greencheck & \greencheck & \greencheck & \greencheck  \\
\hline
\multirow{2}{*}{Visual Feedback} & \multirow{2}{*}{Immersive} & \multirow{2}{*}{Stereoscopic} & Spatial- & High- \\ 
 &  & & unrestricted & fidelity \\\hline
GUI Monitor & \redcross & \redcross & \greencheck & \greencheck \\
Third-person perspective & \redcross & \greencheck & \redcross & \greencheck \\
First-person perspective & \greencheck & \greencheck & \redcross & \greencheck \\
3D Reconstruction & \greencheck & \greencheck & \greencheck & \redcross \\
Stereo Stream & \greencheck & \greencheck & \greencheck & \greencheck \\
\bottomrule

\end{tabular}
\end{center}
\vspace{-6mm}
\end{table}

\vspace{-3mm}
\subsection{Retargeting Module in Teleoperation}
The retargeting module maps measured human motions to robot control commands. Joint-space mapping \cite{zhao2023learning, schwarz2023robust} transmits the joint configurations of the leader arms to the follower manipulators, which requires similar morphologies between leaders and followers. Task-space mapping \cite{laghi2018shared, iyer2024open, ding2024bunny, rouxel2022multicontact, wen2023collaborative, ajoudani2012tele, penco2019multimode, ozdamar2022shared, yang2024ace} widely measures human hand poses as motion references for robot end-effectors, supporting cross-morphological teleoperation. Later, advanced control strategies can be integrated to solve low-level commands \cite{kanoun2011kinematic}. Depending on the selection of the operator's motion reference frames, it can be further classified into natural mode and relative mode. The natural mode \cite{laghi2018shared, ding2024bunny} defines fixed base frames (e.g., shoulders) and computes the hand poses based on these frames. In contrast, the relative mode incorporates a trigger to enable/disable retargeting, setting the base frames to the last-step hand poses. Control commands are updated only when enabling retargeting.
Previous work with natural mode \cite{ding2024bunny, cheng2024open} tuned a fixed scaling factor for position mapping while leaving more challenging orientation mapping as identical alignment. The posture analogy is intuitive for operators to view the robot as their kinematic avatar, which facilitates high efficiency. However, improper factors and overlooking of orientation mismatch sacrifice physical ergonomics.
Some systems use relative mode to relieve the problem at a cost of efficiency, which allows operators to adjust hand postures during pauses between disabling and enabling \cite{ozdamar2022shared}. In CaFe-TeleVision, we propose a coarse-to-fine retargeting mechanism that jointly optimizes efficiency and physical ergonomics.

\subsection{Perception Module in Teleoperation}
Prior research has shown that human vision systems rely on a combination of visual cues to perceive (e.g., depth) and then cognitive processing, including proprioceptive cues, monocular cues, binocular cues, and dynamic cues \cite{el2019survey}. To enhance immersive feelings, Cheng et al. \cite{cheng2024open} streamed stereo video from a robot's eye camera to an HMD to reproduce binocular visual cues. However, limited dynamic cues (motion parallaxes) also hamper cognitive ergonomics, restricted by occlusion-induced blind spots. NimbRo system \cite{schwarz2023robust} mitigated it by using a $6$-DoF neck arm, which allows large-range rotational and translational changes of observation viewpoints. Despite the performance, this superiority comes at the expense of increased system complexity and higher costs. 
Recent approaches based on 3D reconstruction \cite{audonnet2024immertwin, patil2024radiance} offer controllable scenes that support immersive and active operator observation. However, these methods face a fundamental trade-off between high fidelity and computational efficiency. 
Alternatively, multi-view visualization complements visual cues via increased viewpoints, such as close-up observations. Unlike static multi-view layout, situated visualization as an emerging technique highlights displaying data spatially coupled with its context, which is conducive to cognitive processing \cite{wen2022effects}.
In CaFe-TeleVision, we use two wrist cameras and integrate situated visualization \cite{bressa2021s, white2009interaction} to enhance cognitive ergonomics.

\begin{figure}[t]
    \centering
    \includegraphics[width=0.92\columnwidth]{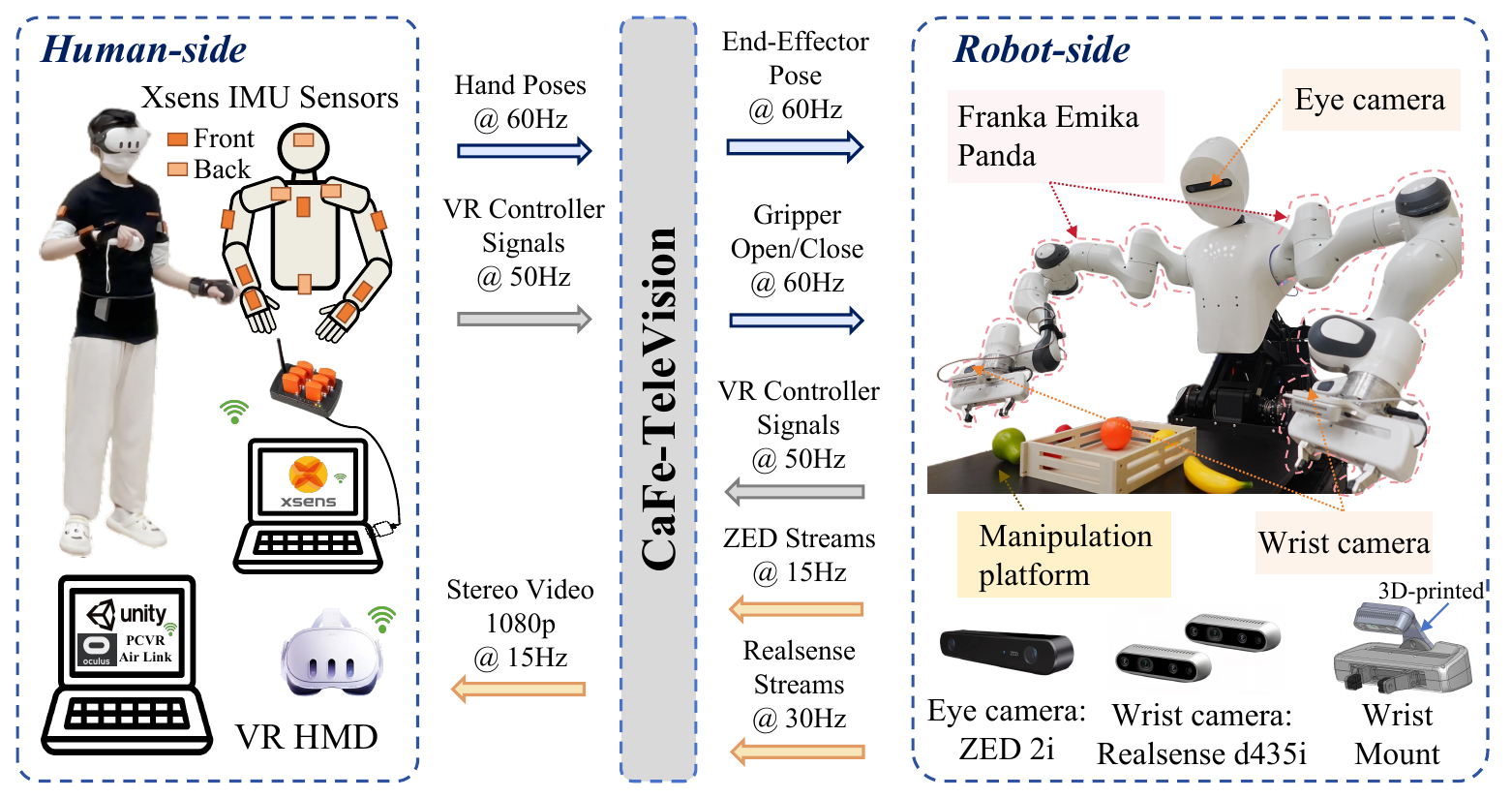}
\caption{Workflow of \textbf{CaFe-TeleVision} teleoperation system. 
The retargeting flow ($\Rightarrow$) supports real-time control, retargeting human motions measured by Xsens and VR controllers to robot commands. The perception flow ($\Leftarrow$) streams multi-view stereo video from robot side to a VR head-mounted display (HMD) through a GPU-accelerated Unity application, ensuring high-resolution, real-time visual feedback.}
\label{fig: workflow}
\vspace{-5mm}
\end{figure}

\begin{figure}[t]
    \centering
    \includegraphics[width=0.93\columnwidth]{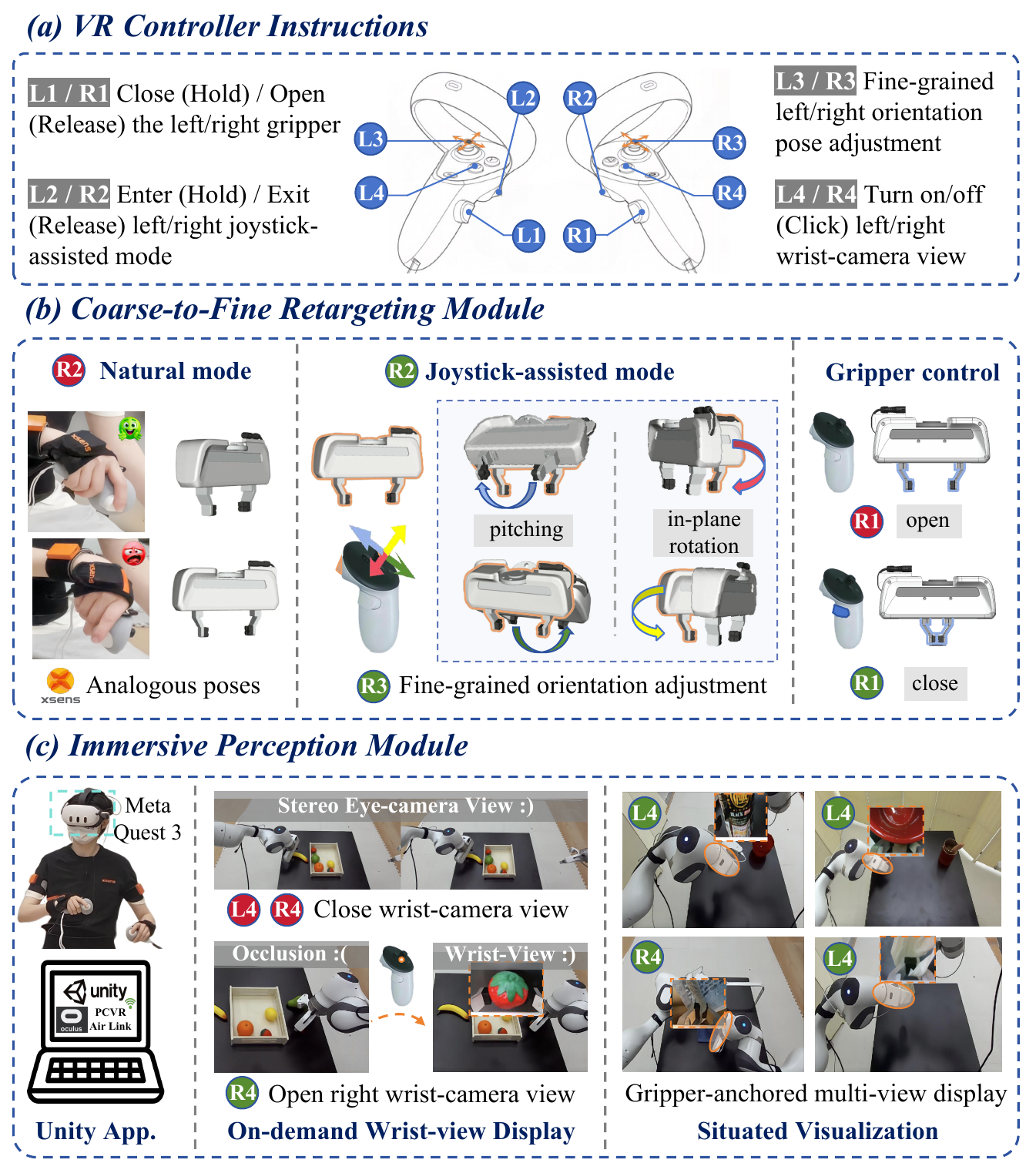}
\caption{Framework of \textbf{CaFe-TeleVision} teleoperation system. At its core, (a) depicts the functionality of VR controller interface, utilized for assisting teleoperation. (b) shows our coarse-to-fine retargeting module with natural mode for efficient motions and joystick-assisted mode for enhanced physical ergonomics via orientation refinements. (c) illustrates our immersive perception module with on-demand situated visualization technique to improve cognitive ergonomics during multi-processing. Button states: inactivate (%
\protect\begin{tikzpicture}
    \protect\filldraw[fill=buttonred, draw=white, line width=0pt] (0,0) circle (0.1cm);
\protect\end{tikzpicture}) and active (\protect\begin{tikzpicture}
    \protect\filldraw[fill=buttongreen, draw=white, line width=0pt] (0,0) circle (0.1cm);
\protect\end{tikzpicture}).
}
\label{fig: framework}
\vspace{-5mm}
\end{figure}

\vspace{-1mm}
\section{Teleoperation System}
\label{sec: method}

\subsection{Overview}
CaFe-TeleVision is a practical and easily deployable VR-based teleoperation system, built on a humanoid collaborative robot. As shown in Fig.~\ref{fig: workflow}, the workflow forms a retargeting-perception loop between human and robot sides. 
Fig.~\ref{fig: framework} presents core designs in our system, including a coarse-to-fine retargeting module and an immersive perception module, along with VR controller interface utilized in them to enhance ergonomics. The complete system implementation is available online on the project's website.

In retargeting flow, we use $11$ Xsens sensors \cite{roetenberg2009xsens} to capture high-accuracy hand poses in SE(3) relative to shoulder frame, which directly reflects the target transformation references in human arm workspace.
The poses are transmitted via UDP packets at $60~\text{Hz}$. Simultaneously, VR controllers signals are input at $50~\text{Hz}$. After retargeting, the output $7$-DoF commands per arm are published at $60~\text{Hz}$ via ROS, then converted into $7$-DoF joint torques through a model-predictive impedance controller \cite{chen2025unified} for Franka arm, with gripper open/close command. The controller operates at $500~\text{Hz}$, integrated with constraints on position, speed, control, and contact torques, which guarantees stable, smooth, and real-time control.

In perception flow, a ZED 2i stereo camera is mounted as the robot's eye, providing a global view of the manipulation platform. 
Two RealSense D435i cameras are mounted on the flanges with 3D-printed adapters. These video streams are processed in Unity application running on a GPU-accelerated laptop, supporting 1080p@$15~\text{Hz}$ real-time visual feedback.

\vspace{-3mm}
\subsection{Coarse-to-Fine Retargeting Module}

The retargeting module maps operator hand poses to robot commands. However, human-robot morphological and kinematic discrepancies inherently cause the workspace mismatch. For example, the human wrist has a limited range of motion, making certain rotations difficult or physically uncomfortable for the operator. In contrast, the robot with a different structural design may handle those motions more easily. To this end, we propose a coarse-to-fine mechanism with two modes: the natural mode for efficient but coarse-grained motions and the joystick-assisted mode for ergonomic and fine-grained adjustments. The dual modes complement each other and jointly optimize efficiency and physical ergonomics.

\subsubsection{Natural mode} The mode follows the pattern of scaling position and aligning orientation to map human wrist motions into analogous target $6$-DoF poses for end-effectors. 
Specifically, the workspaces of both hands are modeled as separate spheres, where the origin is on the operator's shoulder and the radius is arm length, denoted as $r_H$. The intersection area depends on the origin distance of spheres, defined as $d_H$. To achieve full coverage and maintain physical ergonomics in bimanual manipulation, we calculate the scaling factor $\boldsymbol{s}$ for each axis based on the mismatch in radii and origin distances:
\begin{equation}
\boldsymbol{s}=\begin{bmatrix} \frac{r_C}{r_H} & \max(\frac{r_C}{r_H}, \frac{d_C}{d_H}) & \frac{r_C}{r_H}\end{bmatrix}^T,
\end{equation}
where $r_C$ is the workspace radius of Franka, and $d_C$ is its origin distance. Unlike anthropomorphic arms, the origin of Franka lies in the second joint. We thereby transform the retargeted position from the origin to the tilted arm base via $\prescript{i}{}{\boldsymbol{T}}^{ab}_{w}$, where $i\in\{\text{L}, \text{R}\}$ denotes the arm side. The retargeted position $\prescript{i}{}{\boldsymbol{p}}^{ab}_{ee}(t)$ for the end-effector is computed as:
\begin{equation}
\label{retargeting:p}
\prescript{i}{}{\tilde{\boldsymbol{p}}}^{ab}_{ee}(t) = \prescript{i}{}{\boldsymbol{T}}^{ab}_{w}\prescript{i}{}{\tilde{\boldsymbol{p}}}^{w}_{ee}(t), \ \ \ \prescript{i}{}{\boldsymbol{p}}^{w}_{ee}(t)=\boldsymbol{s}\circ\prescript{i}{}{\boldsymbol{p}}^{w}_{hand}(t),
\end{equation}
where human hand position $\prescript{i}{}{\boldsymbol{p}}^{w}_{hand}(t)$ based on the shoulder frame is obtained from Xsens in real time, $\circ$ is Hadamard product, and $\tilde{\boldsymbol{p}}$ refers to homogeneous coordinate format.

We align the $3$-DoF orientation via $\prescript{i}{}{\boldsymbol{R}}_{ee}^{hand}$, identically mapping hand to the approaching direction, the rotation in ulnar/radial deviation to in-plane rotation of the grippers and extension/flexion to pitching. Then, the retargeted orientation $\prescript{i}{}{\boldsymbol{R}_{ee}^{ab}(t)}$ related to the arm base frame is defined as:
\begin{equation}
\label{eq: R}
\prescript{i}{}{\boldsymbol{R}_{ee}^{ab}(t)}=\prescript{i}{}{\boldsymbol{R}}_{w}^{ab}\prescript{i}{}{\boldsymbol{R}}_{hand}^{w}(t)\prescript{i}{}{\boldsymbol{R}}_{ee}^{hand},
\end{equation}
where $\prescript{i}{}{\boldsymbol{R}}_{hand}^{w}(t)$ is the orientation of hand in real time. Benefiting from intuitiveness, this mode supports coarse-grained but high-amplitude motions, ensuring high efficiency.

\subsubsection{Joystick-assisted mode} When facing non-ergonomic risk in natural mode, operators can seamlessly switch to joystick-assisted mode by holding L2/R2, as exemplified in Fig.~\ref{fig: framework} (b). Fine-grained pose adjustments can be applied using $2$-DoF thumbsticks, one for in-plane rotation and the other for pitching. And the hypothesis for fixing the approaching direction is that humans have greater freedom to adjust their wrists to feasible directions in natural mode. Note that position mapping still follows the natural mode.

When holding the R2 button, the orientation of right gripper stops aligning with Eq.~\ref{eq: R}. Instead, it maintains a continuous $2$-DoF rotation following the scrolling of R3. Formally, let $\prescript{i}{}{u_1}, \prescript{i}{}{u_2}\in [-1, 1]$ denote the normalized $2$-DoF signal from the $i$-side thumbstick, where $\prescript{i}{}{u_1}$ denotes horizontal deflection and $\prescript{i}{}{u_2}$ is vertical deflection. To enhance action-perception consistency, we implement a transparent spatial mapping: horizontal inputs (left-right) control gripper's in-plane rotation and vertical inputs (up-down) adjust its pitching. Then, $\prescript{i}{}{u_1}, \prescript{i}{}{u_2}$ are retargeted to incremental angular change, defined as:
\begin{equation}
\prescript{i}{}{\theta_1}=s_1\prescript{i}{}{u_1}, \ \ \ \prescript{i}{}{\theta_2}=s_2\prescript{i}{}{u_2},
\end{equation}
where $\prescript{i}{}{\theta_1}$, $\prescript{i}{}{\theta_2}$ denote in-plane angle and pitch angle, respectively. $s_1$ and $s_2$ are rotational scaling factors. Assuming $i$-side joystick-assisted mode engaged at $t_0$, $\prescript{i}{}{\boldsymbol{R}}_{ee}^{ab}(t_0)$ is recorded as the initial orientation, then updated as:
\begin{equation}
\label{retargeting:fine}
\prescript{i}{}{\boldsymbol{R}_{ee}^{ab}(t)}\leftarrow\prescript{i}{}{\boldsymbol{R}_{ee}^{ab}(t)}\Delta\boldsymbol{R}(\prescript{i}{}{\theta_2(t)})\Delta\boldsymbol{R}(\prescript{i}{}{\theta_1(t)}),
\end{equation}
where $\Delta\boldsymbol{R}(\prescript{i}{}{\theta_1(t)})$ and $\Delta\boldsymbol{R}(\prescript{i}{}{\theta_2(t)})$ (see Eq.~\ref{retargeting:delta}) are rotation for in-plane direction and pitching, respectively. 
\begin{equation}
\label{retargeting:delta}
\begin{split}
\Delta\boldsymbol{R}(\prescript{i}{}{\theta_1(t)})=
\begin{bmatrix}
\cos(\prescript{i}{}{\theta_1(t)}) & -\sin(\prescript{i}{}{\theta_1(t)}) & 0 \\
\sin(\prescript{i}{}{\theta_1(t)}) & \cos(\prescript{i}{}{\theta_1(t)}) & 0 \\
0 & 0 & 1
\end{bmatrix},\\
\Delta\boldsymbol{R}(\prescript{i}{}{\theta_2(t)})=
\begin{bmatrix}
\cos(\prescript{i}{}{\theta_2(t)}) & 0 & \sin(\prescript{i}{}{\theta_2(t)}) \\
0 & 1 & 0 \\
-\sin(\prescript{i}{}{\theta_2(t)}) & 0 & \cos(\prescript{i}{}{\theta_2(t)})
\end{bmatrix}\ 
\end{split}
\end{equation}

The integration of the joystick-assisted mode in retargeting module allows operators to maintain ergonomic postures while gradually adjusting to feasible poses. 
When L2/R2 is released, $\prescript{i}{}{\boldsymbol{R}_{ee}^{ab}(t)}$ is interpolated to real-time human hand poses via SLERP algorithm, smoothly rotating gripper towards targets at a constant angular velocity. The interpolation automatically terminates if the rotation distance between gripper’s current orientation and the hand orientation is below the threshold. Then, the identical alignment in natural mode is recovered immediately.

\begin{figure}[t]
    \centering
    \includegraphics[width=0.96\columnwidth]{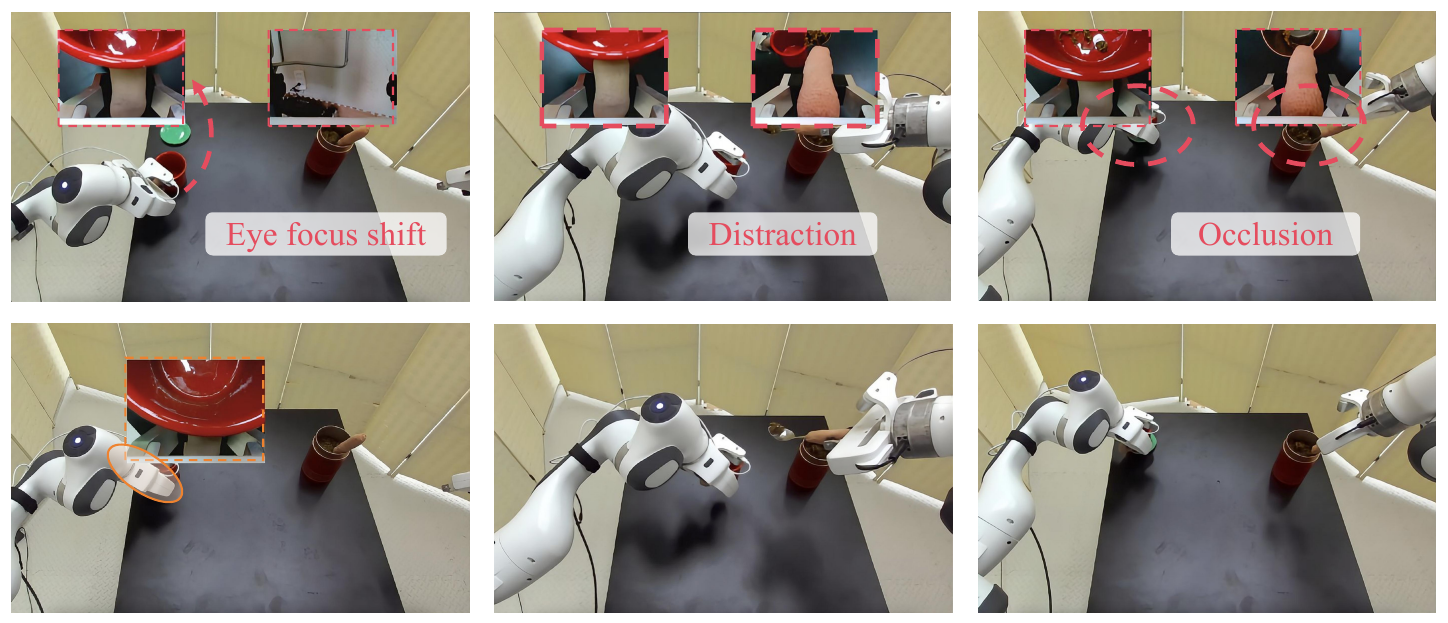}
\caption{Comparison of multi-view display between ``always on" static layout (top) and on-demand situated visualization (bottom). The bottom pattern (ours) leads to low cognitive load, in terms of eye focus shift, distraction, and occlusion.}
\label{fig: on_demand_visibility}

\end{figure}

\begin{figure*}[t]
    \centering
    \includegraphics[width=1.89\columnwidth]{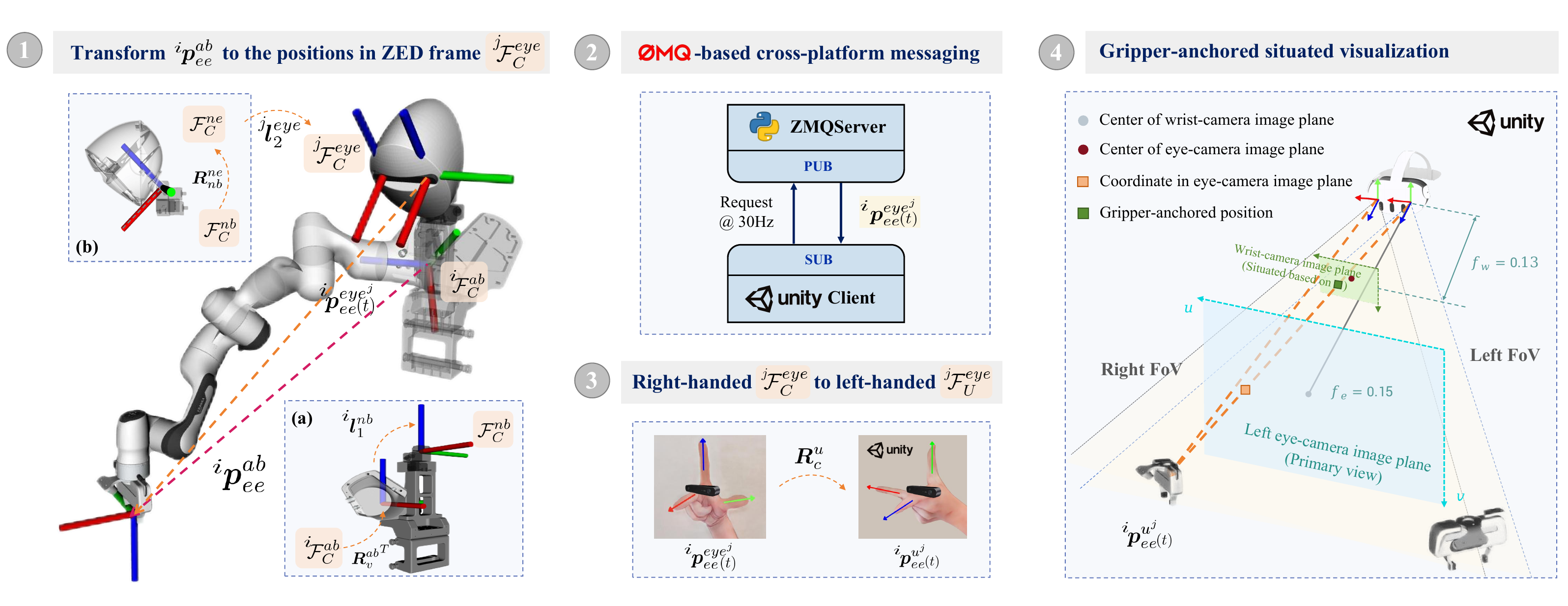}
\caption{Technical pipeline of gripper-anchored situated visualization. Specifically, step 1 calculates the gripper-to-eye translation. Step 2 messages the translation to the unity client via ZeroMQ. Step 3 transforms the translation based on right-handed rule to the one in left-handed rule. Step 4 situates the wrist-camera view near the gripper-anchored position.}
\label{fig: situated_visualization}
\vspace{-3mm}
\end{figure*}

\vspace{-3mm}
\subsection{Immersive Perception Module}
Recent work leverages VR HMDs to reproduce binocular vision \cite{cheng2024open}. Despite the progress, these systems remain unsatisfactory due to inadequate visual cues such as dynamical cues \cite{el2019survey}.  Building upon this evidence, we maintain stereoscopic streaming from the eye camera as a high-fidelity primary display, and complement other visual cues via two wrist cameras. As shown in Fig.~\ref{fig: framework} (c), CaFe-TeleVision features two critical designs: on-demand display and situated visualization.

\subsubsection{On-demand wrist-view display}
Many multi-view VR applications non-stop render all views during runtime, regardless of whether a view is useful for user decisions and behaviors, as shown in Fig.~\ref{fig: on_demand_visibility}. This ``always-on'' rendering strategy causes visual distraction and occlusion. Hence, we design an on-demand display strategy that allows operators to toggle the visibility of wrist-camera views using buttons.

A practical scenario is shown in Fig.~\ref{fig: framework} (c). When grasping a banana, the eye-camera view offers operators adequate visual information to locate and pick it without additional views. However, the other case is for a strawberry. Occlusion caused by gripper itself obscures successful grasping. In this circumstance, operators typically activate the right wrist-camera view via R4 to obtain close-up observations. This on-demand pattern improves depth perception and detail awareness, leading to a high grasping success rate and a low cognitive load.

\subsubsection{Gripper-anchored situated visualization}
Situated visualization \cite{white2009interaction} is an emerging technique that renders graphical elements within their spatially relevant contexts (i.e., situatedness). Compared with static layout, high situatedness is shown to support understanding and decision-making \cite{wen2022effects}. As such, a gripper-anchored situatedness is implemented in Unity, geometrically aligning the $i$-side wrist-camera view with $i$-side gripper in the eye-camera view. This gripper-anchored pattern is based on the observation that operators always cry for referring additional visual cues when performing contact-rich manipulation, with the gaze naturally focusing on the gripper. So the complementary cues are presented close to the gripper to reduce attention shift, thereby improving cognitive ergonomics in multi-view processing (see Fig.~\ref{fig: on_demand_visibility}).

To preserve stereoscopic perception, two paired views are maintained to cast the observations from $i$-side wrist camera.
Formally, let $j\in\{\text{L}, \text{R}\}$ denote the eye side. Fig.~\ref{fig: situated_visualization} illustrates the four-step pipeline of how to situate the $i$-side wrist-camera view for $j$-side eye in the virtual Unity scene, exemplified with $i=R$ and $j=L$. First, the position of the $i$-side gripper in CURI's $j$-side eye-camera frame $\prescript{j}{}{\mathcal{F}}_C^{eye}$ is:
\begin{equation}
\prescript{i}{}{\boldsymbol{p}}^{eye^j}_{ee}(t) = \boldsymbol{R}_{nb}^{ne}({\prescript{i}{}{\boldsymbol{R}}^{ab}_{w}}^T \prescript{i}{}{\boldsymbol{p}}^{ab}_{ee}(t) + \prescript{i}{}{\boldsymbol{l}}^{nb}_1) + \prescript{j}{}{\boldsymbol{l}}_{2}^{eye},
\end{equation}
where $\boldsymbol{R}_{nb}^{ne}$ is the rotation matrix from neck base frame ${\mathcal{F}}^{nb}_{C}$ to neck end frame ${\mathcal{F}}^{ne}_{C}$, 
$\prescript{i}{}{\boldsymbol{l}}^{nb}_1$ and $\prescript{j}{}{\boldsymbol{l}}_{2}^{eye}$ represent constant translation vectors to frame ${\mathcal{F}^{nb}_C}$ and $\prescript{j}{}{\mathcal{F}}^{eye}_{C}$, respectively.
Second, $\prescript{i}{}{\boldsymbol{p}}^{eye^j}_{ee}(t)$ is messaging to Unity via ZeroMQ at subscribing frequency $30$ Hz. Third, the rotation transformation from real-world frame $\prescript{j}{}{\mathcal{F}}_C^{eye}$ to frame $\prescript{j}{}{\mathcal{F}}^{eye}_{U}$ in Unity is defined by $\boldsymbol{R}_c^u$, so $\prescript{i}{}{\boldsymbol{p}}^{u^j}_{ee}(t)=\boldsymbol{R}_c^u\prescript{i}{}{\boldsymbol{p}}^{eye^j}_{ee}(t)$. Fourth, to overlay the primary view (focal length $f_e$) in Unity, the focal length for the wrist-camera imaging plane is set to $f_w<f_e$. The $i$-side gripper-anchored position \begin{tikzpicture}
    \filldraw[fill=mygreen, draw=mygreenborder, line width=1pt] (0,0) rectangle (0.18cm, 0.18cm);
\end{tikzpicture} for the wrist-camera imaging plane in $j$-side eye is projected as:
\begin{equation}
\prescript{i}{}{\boldsymbol{p}}^{u^j}_{anchor}(t) = \begin{bmatrix} 
\prescript{i}{}{p}^{u^j}_{ee,x}(t)f_w/\prescript{i}{}{p}^{u^j}_{ee,z}(t) \\ 
\prescript{i}{}{p}^{u^j}_{ee,y}(t)f_w/\prescript{i}{}{p}^{u^j}_{ee,z}(t) \\
f_w
\end{bmatrix}
\end{equation}
Finally, the $i$-side wrist-camera view for $j$-side eye is positioned on \begin{tikzpicture}
    \filldraw[fill=myred, draw=myred, line width=0pt] (0,0) circle (0.1cm);
\end{tikzpicture}, with pre-defined in-plane translation and scale related to \begin{tikzpicture}
    \filldraw[fill=mygreen, draw=mygreenborder, line width=1pt] (0,0) rectangle (0.18cm, 0.18cm);
\end{tikzpicture} to alleviate occlusion between views.

\section{Experiments}
\label{Experiments}
We conduct a user study to evaluate the performance of teleoperation systems. This section introduces the experimental setup, followed by evaluation metrics concerning success rate, efficiency, and ergonomics. Finally, we summarize quantitative and qualitative results, supported by statistical analysis.

\vspace{-6mm}
\subsection{Experimental Setup}
\subsubsection{Participant Profile} For user study, $24$ participants ($16$ male and $8$ female) were recruited through university-wide open invitation, aged between $19$ and $33$ years, with an average age of $24.96$ and a standard deviation (SD) of $3.34$. Their prior teleoperation experience and VR experience were assessed using a 5-point Likert scale questionnaire ($1$ for ``Expert" and $5$ for ``Novice"), averaging scores of $3.2$ for teleoperation (SD=$1.58$) and $3.4$ for VR experience (SD=$1.32$), respectively.

\subsubsection{Task Suite}
\begin{figure}[t]
    \centering
    \includegraphics[width=0.96\columnwidth]{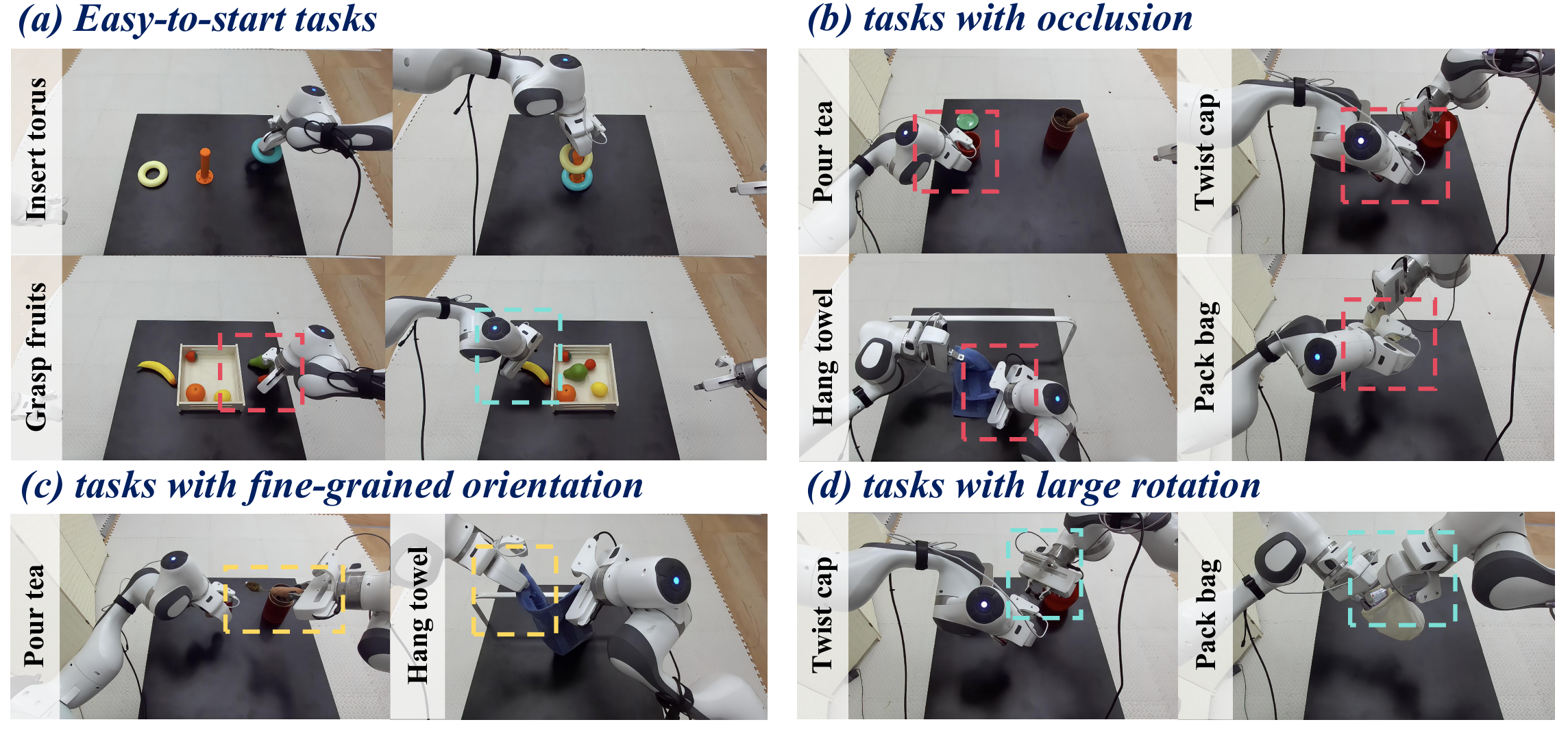}
\caption{Illustration of key phases and characteristics of six manipulation tasks. (a) shows two easy-to-start tasks for novices. (b)-(d) denotes tasks facing challenges in occlusion, fine-grained orientation control, and large rotation, respectively. Each challenge is highlighted with a distinct colored box.}
\label{fig: tasks}
\vspace{-5mm}
\end{figure}

To challenge the capabilities of teleoperation systems, we elaborately designed six tasks, including \textit{insert torus}, \textit{grasp fruits}, \textit{pour tea}, \textit{twist cap}, \textit{hang towel}, \textit{pack bag}. Fig.~\ref{fig: tasks} shows critical phases and the characteristics for each task. Specifically, challenges involve: fine-grained orientation control, handling a large range of rotation outside of the ergonomic joint range of operators, and visual occlusion. All characteristics set an effective testbed to evaluate the strengths of our system. More details are available in the supplementary video\footnote{CaFe-TeleVision project webpage: \url{https://clover-cuhk.github.io/cafe_television/}} (e.g., the prescribed time limit for each task).

\subsubsection{Baseline Systems}
To evaluate the effectiveness of CaFe-TeleVision, we compare it with three representative systems. These baselines adopt standard retargeting and perception schemes, typical in state-of-the-art teleoperation systems. For brevity, we refer to N as natural mode, R as relative mode, S as stereo visual feedback, and SL as static multi-view layout.
\begin{itemize}
\item \textbf{N-S}: Adopt natural mode for retargeting and only streams the stereo video from the eye camera, without the wrist-camera views, similar to Bunny-VisonPro \cite{ding2024bunny}.
\item \textbf{CaFe-SL}: A variant of CaFe-TeleVision that renders multiple views in the static layout instead of on-demand situated visualization technique.
\item \textbf{R-TeleVision}: A variant of CaFe-TeleVision that replaces coarse-to-fine mechanism with relative mode for retargeting. Each time, retargeting is toggled by holding L2/R2 button, and the relative pose transformation depends on the initial pose when pressing the buttons.
\end{itemize}

\subsubsection{Experimental Design}
Pilot studies indicate that the difficulty of several tasks (such as \textit{pour tea} and \textit{pack bag}) exceeds what novices can successfully complete within a short practice period (e.g., 5 minutes). Incorporating excess low-quality results overwhelms the statistical reliability of system comparison. Towards this, the experiments are designed into two parts. First, two easy-to-start tasks (i.e., \textit{insert torus} and \textit{grasp fruits}) are selected for all participants to perform. The results from this wide demographic enhance the comparison analysis with high statistical power. Second, three expert participants evaluate all six tasks, where experts denote those who self-rated both teleoperation and VR experience as 1 (indicating ``Expert"). Benefiting from low proficiency-related variability, failure case analysis on the results is thus expected to reveal the inherent capabilities and limitations of each system in handling challenging scenarios. The experimental durations for each non-expert and expert are within 3 hours and 12 hours, respectively. The compensation depends on the working hours, at a rate of HK \$64 per hour.

All experiments employ a within-subjects design, where each participant evaluates all four systems across testing tasks, performing five trials with each system per task to enhance measurement reliability. In total, each participant executes $20$ trials per task. To alleviate learning and order effects, we implemented rigorous order control throughout the experiments. Specifically, a partial counterbalancing design with full permutations of systems was adopted for the all-participant experiment. In detail, a complete counterbalancing of task order was employed, evenly assigning two task order sequences to all participants. We then fully covered all possible system permutations among four systems and assigned a unique system order to each participant. For the expert experiment, six tasks are grouped by bimanual coordination levels (see Fig.~\ref{fig: overview}), and the group order assigned to each expert is based on Latin Square counterbalancing. The task order within groups and the system order were randomized.

\begin{table}[t]
\caption{Quantitative teleoperation performance on \textit{insert torus} and \textit{grasp fruits} task, in terms of \textit{success rate (SR)} and \textit{completion time (Time)} in seconds. The results are average over 120 trials from all participants. Note: \textit{TeleVision} is abbreviated as \textit{Tele}.}

\label{tab: 2}
\begin{center}
\setlength{\tabcolsep}{1.3pt}
\begin{tabular}{ccccccccccccc}
\toprule
\multirow{2}{*}{\textbf{Method}} & \multicolumn{2}{c}{\textbf{Insert torus}} & \multicolumn{2}{c}{\textbf{Grasp fruits}} \\
\cmidrule(lr){2-3}\cmidrule(lr){4-5}
 & \textbf{SR} $\uparrow$ & \textbf{Time} $\downarrow$ & \textbf{SR} $\uparrow$ & \textbf{Time} $\downarrow$ \\
\midrule
N-S & 96.81\%$\pm$17.67\% & 36.97$\pm$12.08 & 70.00\%$\pm$46.08\% & 79.99$\pm$35.70 \\
CaFe-SL & 94.68\%$\pm$22.56\% & 39.52$\pm$14.84 & 73.33\%$\pm$41.47\% & 80.83$\pm$37.03   \\
R-Tele & 81.92\%$\pm$38.40\% & 49.04$\pm$18.36 & 54.44\%$\pm$50.08\% & 100.79$\pm$39.19 \\ 
\addlinespace[0.5pt]
\hline
CaFe-Tele & \textbf{97.87\%}$\pm$\textbf{14.51\%} & \textbf{33.90}$\pm$\textbf{11.29} & \textbf{78.89\%}$\pm$\textbf{41.04\%} & \textbf{72.44}$\pm$\textbf{34.89}  \\ 
\bottomrule
\end{tabular}
\end{center}
\vspace{-3mm}
\end{table}

\begin{table}[t]
\caption{Post-hoc Nemenyi tests of \textit{completion time} metric on \textit{insert torus} and \textit{grasp fruits} tasks. Significant pairs ($p < 0.05$) are listed with p-values to six decimal places.}
\label{tab: 3}
\centering
\setlength{\tabcolsep}{6pt} 
\renewcommand{\arraystretch}{1.05} 
\begin{tabular}{@{}ccc@{}}
\toprule
\textbf{Task} & \textbf{Significant pair} & \textbf{\textit{p}-value} \\
\midrule
Insert torus & CaFe-TeleVision vs. R-TeleVision & 0.000003 \\
\addlinespace[0.5pt]
\hline
\multirow{3}{*}{Grasp fruits} 
    & CaFe-TeleVision vs. N-S & 0.020126 \\
    & CaFe-TeleVision vs. R-TeleVision & 0.000002 \\
    & CaFe-SL vs. R-TeleVision & 0.020126 \\
\bottomrule
\end{tabular}
\vspace{-3mm}
\end{table}


\begin{table*}[t]
\caption{Statistical analysis on six-dimensional NASA-TLX scores from all participants, using Friedman tests with post-hoc Nemenyi tests for pairwise evaluation. Assessments include mental demand (MD), physical demand (PD), temporal demand (TD), performance, effort, and frustration. P-values are reported to four decimal places, with significance ($p<0.05$) bolded.}

\label{tab: 4}
\vspace{-2.5mm}
\begin{center}

\begin{tabular}{c|c|ccccccc}
\toprule
\multicolumn{2}{c}{} & \multicolumn{6}{c}{\textbf{Six-dimensional task load index}} & \multirow{2}{*}{\textbf{Mean}} \\
\cmidrule(lr){3-8}
\multicolumn{2}{c}{} & \textbf{MD} & \textbf{PD} & \textbf{TD} & \textbf{Performance} & \textbf{Effort} & \textbf{Frustration} &   \\
\midrule
\multirow{2}{*}{\makecell{Friedman\\test}} & $\chi^2(3)$ & 11.00 & 16.16 & 15.19 & 14.49 & 14.20 & 13.20 & 17.93 \\
& \textit{p}-value & \textbf{0.0117} & \textbf{0.0010} & \textbf{0.0012} & \textbf{0.0023} & \textbf{0.0026} & \textbf{0.0042} & \textbf{0.0005} \\
\hline
\multirow{6}{*}{\makecell{Post-hoc\\system\\vs.\\system\\pair}} & CaFe-TeleVision vs. N-S & 0.3709 & \textbf{0.0037} & \textbf{0.0179} & \textbf{0.0015} & \textbf{0.0305} & \textbf{0.9999} & \textbf{0.0024}  \\
& CaFe-TeleVision vs. CaFe-SL & 0.0911 & 0.6675 & 0.1371 & 0.3064 & 0.0789 & 0.8903 & \textbf{0.0215}  \\
& CaFe-TeleVision vs. R-TeleVision & \textbf{0.0149} & 0.9952 & \textbf{0.0024} & 0.1762 & \textbf{0.0069} & 0.0911 & \textbf{0.0015} \\ 
& N-S vs. CaFe-SL & 0.8903 & 0.1049 & 0.8650 & 0.2227 & 0.9842 & 0.8650 & 0.9130 \\
& N-S vs. R-TeleVision & 0.5154 & \textbf{0.0083} & 0.9328 & 0.3709 & 0.9640 & 0.1049 & 0.9994 \\
& CaFe-SL vs. R-TeleVision & 0.9130 & 0.8067 & 0.5154 & 0.9907 & 0.8370 & \textbf{0.0124} & \textbf{0.8650} \\
\bottomrule
\end{tabular}

\end{center}
\vspace{-4mm}
\end{table*}



\begin{table*}[t]
\caption{Quantitative teleoperation performance from expert participants on six challenging tasks, in terms of \textit{success rate (SR)} and \textit{completion time (Time)} in seconds.}
\label{tab: 5}
\vspace{-2.5mm} 
\begin{center}
\begin{tabular}{ccccccccccccc}
\toprule
\multirow{2}{*}{\textbf{Method}} & \multicolumn{2}{c}{\textbf{Insert torus}}  & \multicolumn{2}{c}{\textbf{Grasp fruits}}  & \multicolumn{2}{c}{\textbf{Pour tea}}  & \multicolumn{2}{c}{\textbf{Twist cap}}  & \multicolumn{2}{c}{\textbf{Hang towel}} & \multicolumn{2}{c}{\textbf{Pack bag}} \\
\cmidrule(lr){2-3}\cmidrule(lr){4-5}\cmidrule(lr){6-7}\cmidrule(lr){8-9}\cmidrule(lr){10-11}\cmidrule(lr){12-13}
 & \textbf{SR} $\uparrow$ & \textbf{Time} $\downarrow$ &\textbf{SR} $\uparrow$ & \textbf{Time} $\downarrow$ & \textbf{SR} $\uparrow$ & \textbf{Time} $\downarrow$ & \textbf{SR} $\uparrow$ & \textbf{Time} $\downarrow$ & \textbf{SR} $\uparrow$ & \textbf{Time} $\downarrow$ & \textbf{SR} $\uparrow$ & \textbf{Time} $\downarrow$  \\
\midrule
N-S & \textbf{100\%} & 30.77 & \textbf{100\%} & 57.59 & 46.67\% & 60.38 & 73.33\% & 117.02 & 73.33\% & 51.67 & 33.33\% & 145.36     \\
CaFe-SL & \textbf{100\%} & \textbf{28.01} & \textbf{100\%} & 52.52 & 86.67\% & 66.49 & 100\% & 100.55 & 86.67\% & 54.64 & 100\% & 110.20  \\
R-TeleVision & \textbf{100\%} & 36.42 & \textbf{100\%} & 70.11 & 80\% & 68.81 & 53.33\% & 142.49 & 100\% & 57.55 & 66.67\% & 132.52 \\ 
\hline
CaFe-TeleVision & \textbf{100\%} & 29.31 & \textbf{100\%} & \textbf{46.25} & \textbf{100\%} & \textbf{57.74} & \textbf{100\%} & \textbf{86.53} & \textbf{100\%} & \textbf{48.89} & \textbf{100\%} & \textbf{103.01}  \\ 
\bottomrule
\end{tabular}
\end{center}
\vspace{-7mm}
\end{table*}

\subsubsection{Experimental Procedure}
Participants received an introduction to teleoperation and experimental objectives. The task and system sequence for evaluation were then assigned based on the established order control. Following this, they calibrated the Xsens motion tracking system based on provided instructions and underwent a 10-minute training session to familiarize themselves with operational keypoints of these systems. For each testing task, they were given a 5-minute practice period before officially recording their performance. After that, the quantitative results were measured during the experiments. At the end, participants completed relevant questionnaires to assess their qualitative subjective feelings.


\subsection{Evaluation Metrics}
\subsubsection{Quantitative Measurements}
To quantitatively analyze performance for specific tasks, two metrics are measured during the experiments. (a) \textit{Success Rate (SR)}: the ratio of successful trials to total trials; (b) \textit{Completion Time (Time)}: the average task completion time across all trials. A task trial is deemed a failure under the following conditions. First, it exceeds the prescribed time. Second, for fine-grained tasks, if the tea is split or towel is not hung flat, the trial is also counted as a failure. Trials above twice the time limit are terminated.

\subsubsection{Qualitative Assessments}
To qualitatively assess the subjective feelings regarding system workload and acceptance, two typical questionnaires were conducted after the experiments. (a) \textit{NASA-TLX}: a six-dimensional task load index, with higher values reflecting greater perceived difficulty in experiments. (b) \textit{System Usability Scale (SUS)}: a 10-item instrument designed to assess subjective usability, with higher values showing a stronger user acceptance of the method.

\subsection{All-Participant Performance Analysis}
Table~\ref{tab: 2} presents the quantitative results on two easy-to-start tasks: \textit{insert torus} and \textit{grasp fruits}. Each item is calculated over $120$ trials from all participants, ensuring measurement reliability. Overall, CaFe-Television achieves superior performance, especially in \textit{grasp fruit} task. To evaluate systems statistically, Friedman tests with post-hoc Nemenyi tests were conducted on \textit{completion time} metric separately for each task. Friedman tests confirmed significant differences among four systems, with $\chi^2(3)=25.30$, $p=0.000013$ for \textit{insert torus} and $\chi^2(3)=30.11$, $p=0.000001$ for \textit{grasp fruits}. Post-hoc Nemenyi tests revealed specific significant pairs, as reported in Table~\ref{tab: 3}. We can see: CaFe-TeleVision completion times were significantly lower than those of R-TeleVision ($p < 0.05$), which demonstrates that our system supports more efficient operation compared to relative mode. For \textit{grasp fruits} task, our system established statistical significance with N-S, validating the effectiveness of the coarse-to-fine control mechanism in the scenario requiring large wrist rotation. But no significant difference was found between CaFe-SL and N-S, indicating the control advantage brought by this mechanism is diminished by the side effects from static multi-view perception. This observation thus reflects that our on-demand situated visualization is desired for multi-view processing.

\begin{figure}[t]
    \centering
    \includegraphics[width=0.65\columnwidth]{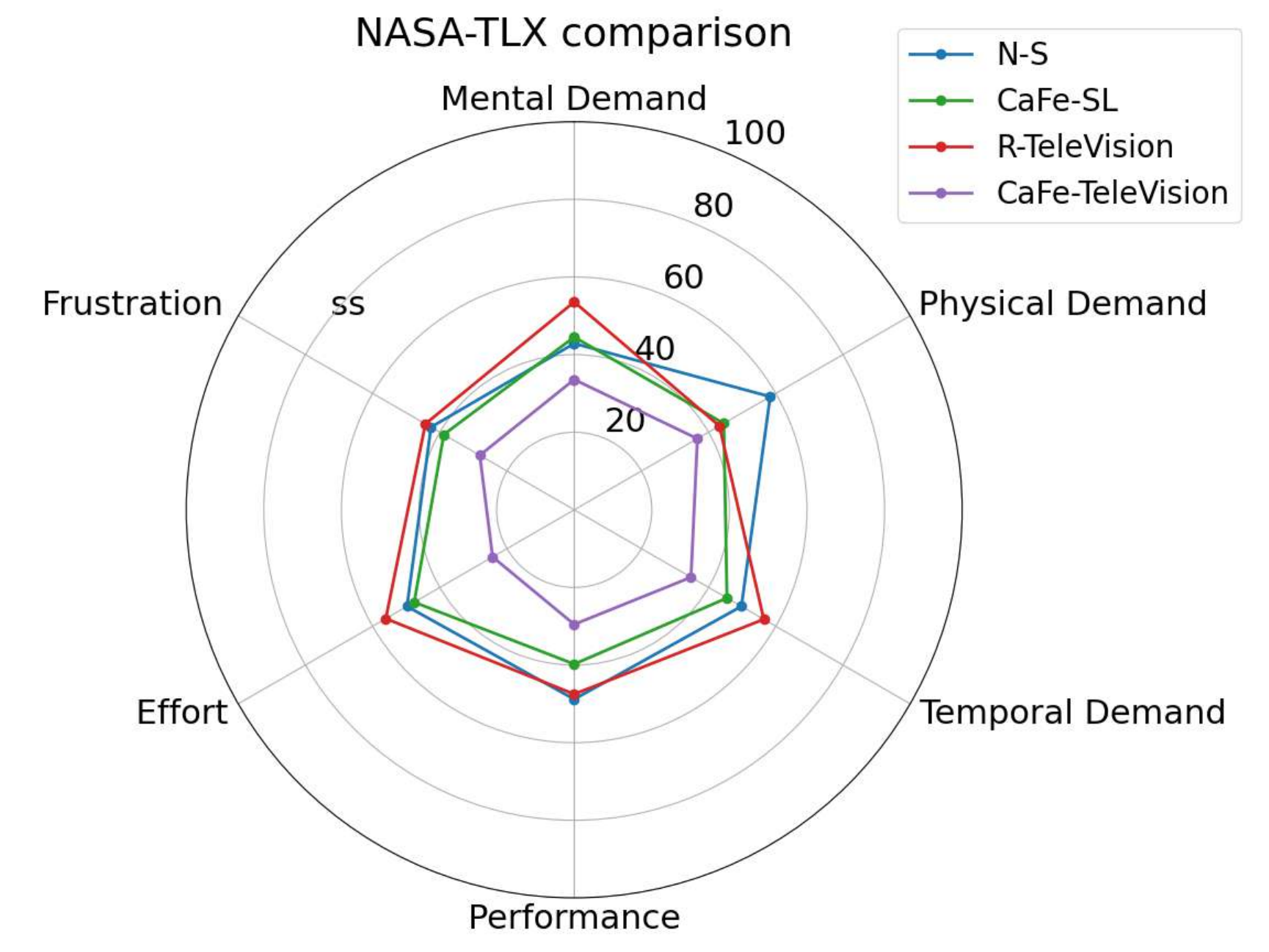}
\caption{NASA-TLX results from all participants on six-dimensional task load scores for four systems. Our system achieves the minimal enclosed area, validating a lower physical and mental workload.}
\label{fig: NASA_TLX}
\vspace{-5mm}
\end{figure}
Fig.~\ref{fig: NASA_TLX} illustrates the six-dimensional workload assessment via NASA-TLX for four systems, with statistical analysis results shown in Table~\ref{tab: 4}. The radar chart demonstrates that CaFe-TeleVision achieves the minimal enclosed area, which confirms that the proposed mechanisms successfully reduce the task load ($p<0.05$, Friedman test). Post-hoc Nemenyi tests report pairwise significances between systems. Notably, the improvement is particularly pronounced in \textit{Effort} and \textit{Performance} dimensions compared to baseline systems, with average improvements of $16.87\%$ and $15.83\%$, respectively. These findings indicate enhanced ergonomics by CaFe-TeleVision, thereby boosting overall performance.

\begin{figure}[t]
    \centering
    \begin{minipage}{0.48\columnwidth}
        \centering
        \includegraphics[width=\linewidth]{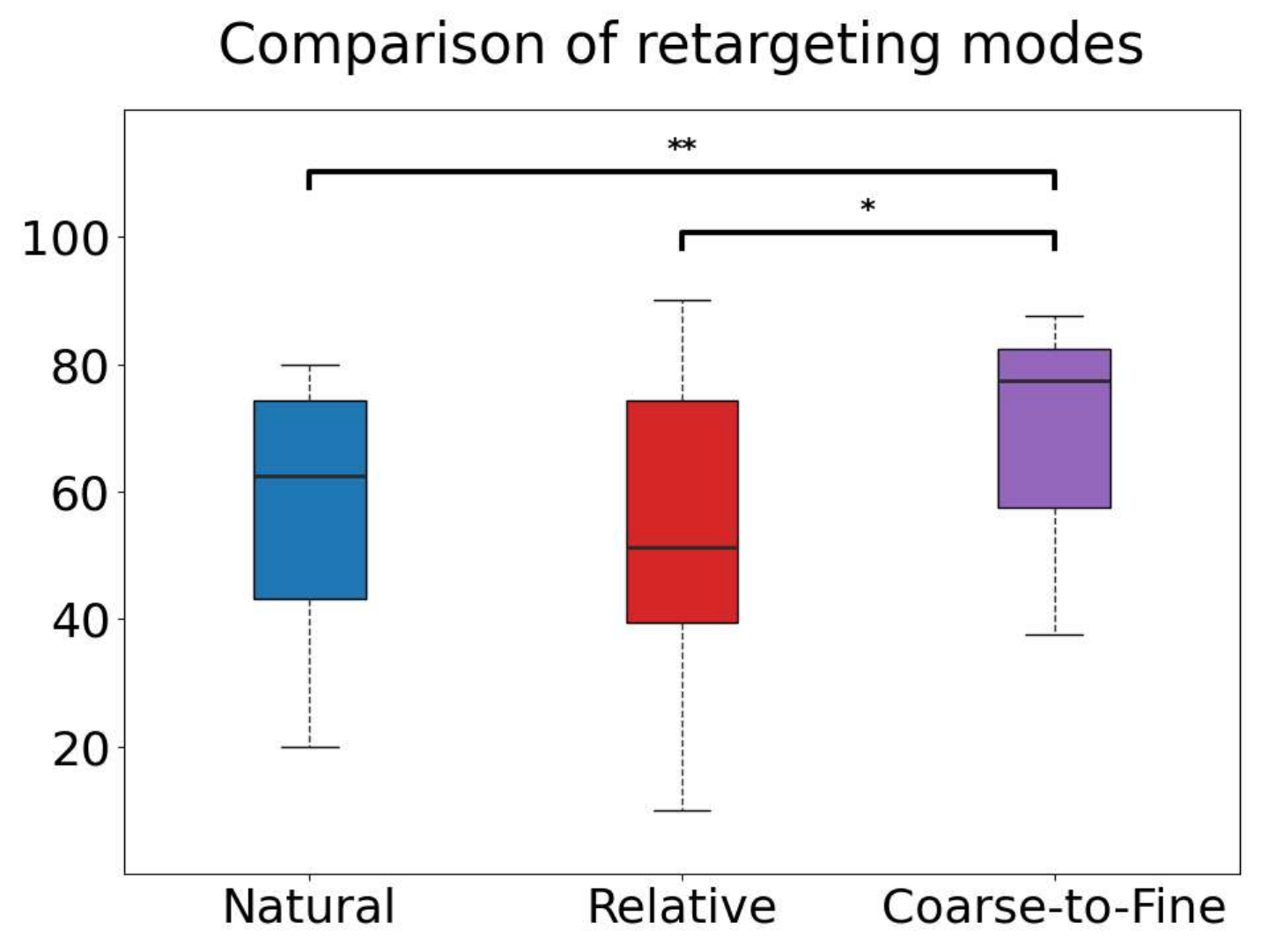}
        \subcaption{Retargeting modes}
        \label{fig:sub_a}
    \end{minipage}
    \begin{minipage}{0.48\columnwidth}
        \centering
        \includegraphics[width=\linewidth]{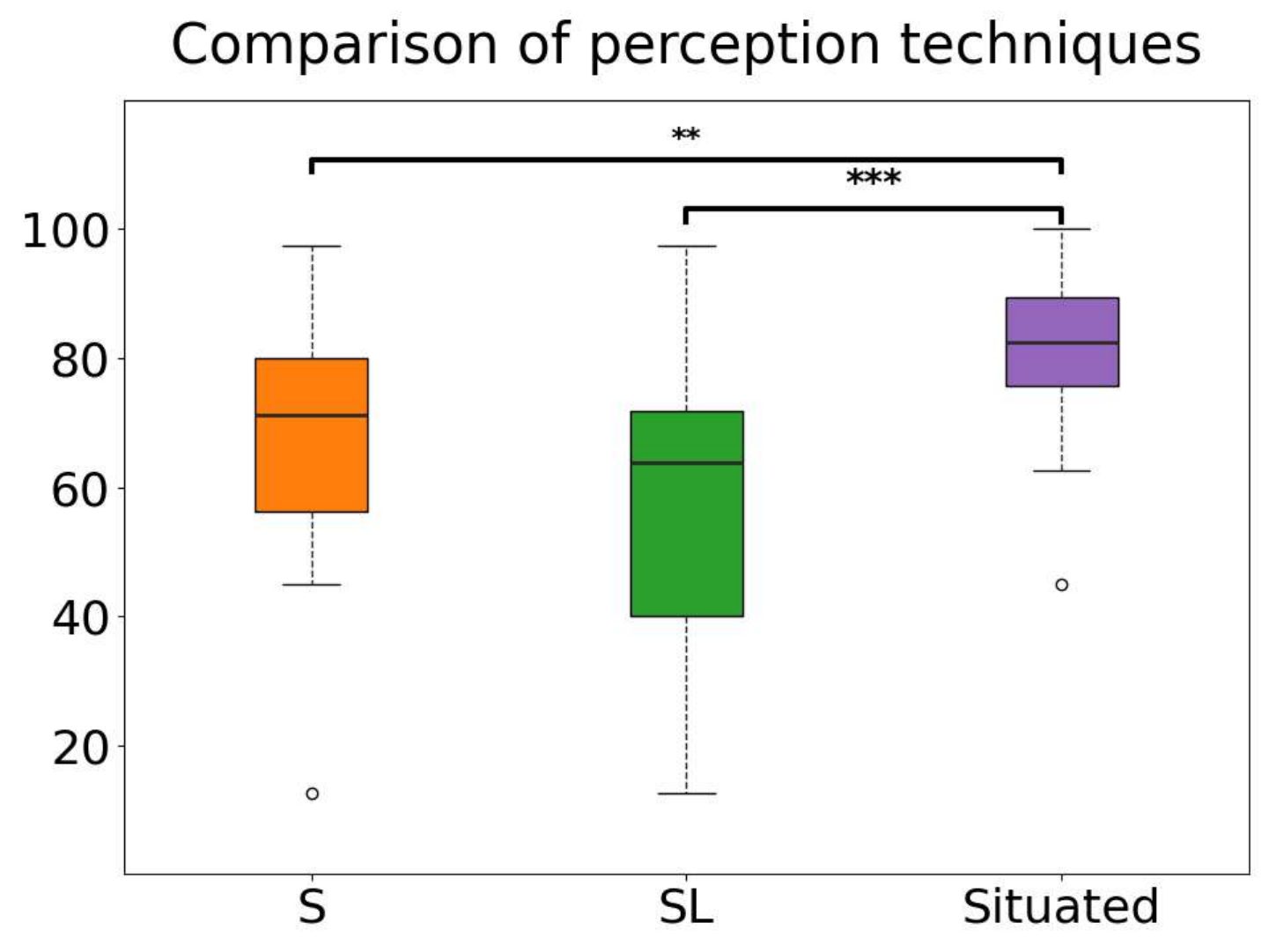}
        \subcaption{Perception techniques}
        \label{fig:sub_b}
    \end{minipage}

\caption{System usability scale (SUS) results from all participants, illustrating higher acceptance of our proposed schemes. (a) compares retargeting modes: natural mode, relative mode, and our coarse-to-fine mechanism. (b) contrasts perception techniques: stereo streaming (S), static multi-view layout (SL), and our on-demand situated visualization (Situated). Significant pairwise differences are mark by asterisks (\text{*}: $p<0.05$, \text{**}: $p<0.01$, \text{***}: $p<0.001$). 
}
\label{fig: SUS}

\vspace{-5mm}
\end{figure}
Fig.~\ref{fig: SUS} (a) and (b) present the System Usability Scale (SUS) results. Friedman tests indicated significant differences among retargeting modes ($\chi^2(3)=11.37$, $p=0.003393$) and perception techniques ($\chi^2(3)=22.07$, $p=0.000016$). Post-hoc Nemenyi tests confirmed pairwise significance, highlighting higher acceptance of our proposed schemes. Statistical evidences show that our coarse-to-fine retargeting was significantly preferred over standard modes ($p<0.05$), especially over natural mode ($p<0.01$). Benefiting from its intuitiveness and seamless switching characteristics, our mechanism jointly optimizes efficiency and physical ergonomics, surpassing relative mode by an average of $25.51\%$. For perceptual feedback,  on-demand situated visualization outperformed static multi-view layouts ($p<0.001$), affirming the cognitive ergonomics benefits of integrating situated visualization techniques in teleoperation. Additionally, we note a higher acceptance (averaging $20.89\%$) of single-view stereo streams over static multi-view displays. We attribute this preference to the detrimental effects of static layouts, such as visual distraction and occlusion, which overshadow the advantages of augmented visual cues during the experiments. This further validates the effectiveness of our perception scheme.

\subsection{Expert Performance Analysis on Challenging Tasks}
Table~\ref{tab: 5} reports the quantitative results from three expert participants on six bimanual teleoperation tasks, facing challenges in orientation and occlusion handling. On average, CaFe-TeleVision consistently improves performance, boosting success rates by up to $28.89$\% and efficiency by $26.81$\%.

Owing to low proficiency-related variability in expert-derived results, 
we can identify the inherent strengths and limitations of each system through failure analysis. For fine-grained tasks, N-S failed in $8$ trials on \textit{pour tea} and $4$ trials on \textit{hang towel} due to difficulties in fine adjustment under natural mapping mode.
In \textit{twist cap} and \textit{pack bag} tasks, operators also struggled to exert large wrist rotation under natural mode due to anatomical limits, leading to uncomfortable postures and timeouts. CaFe-SL failed twice in \textit{pour tea} and \textit{hang towel} tasks, as the fixed wrist views partially occluded operational areas, slowing teleoperation and thus causing timeout failure. R-TeleVision suffers from low efficiency due to the inevitable pauses and non-intuitive orientation retargeting in relative mode, ultimately resulting in severe timeout failures. In contrast, CaFe-TeleVision overcomes these limitations with its coarse-to-fine retargeting and situated visualization technique, leading to superior performance across all complex tasks.

\vspace{-2mm}
\section{Conclusion}

We proposed CaFe-TeleVision, a coarse-to-fine teleoperation system with immersive situated visualization for enhanced ergonomics.
Comprehensive experiments are conducted in the user study, confirming its superior performance and enhanced ergonomics with statistical significance. Moreover, our method bridges teleoperation challenges in a practical and deployable manner, which can be integrated as plug-and-play modules into current VR-based teleoperation systems, yielding immediate and substantial performance improvements.

\textbf{Limitations and Future Work.} First, the transparent spatial mapping in our joystick-assisted mode assumes the gripper approach direction approximately aligns with the viewpoint, which may not generalize to arbitrary configurations. Future work will explore viewpoint-dependent spatial mapping. Second, integration of additional feedback modalities, such as haptic feedback, is desired to enhance cognitive ergonomics, especially in tactile-intensive tasks. Third, developing a collision avoidance mechanism in teleoperation systems is important for ensuring safety. 



\bibliographystyle{IEEEtran}
\bibliography{IEEEabrv,references}

\begin{thebibliography}{10}
\providecommand{\url}[1]{#1}
\csname url@samestyle\endcsname
\providecommand{\newblock}{\relax}
\providecommand{\bibinfo}[2]{#2}
\providecommand{\BIBentrySTDinterwordspacing}{\spaceskip=0pt\relax}
\providecommand{\BIBentryALTinterwordstretchfactor}{4}
\providecommand{\BIBentryALTinterwordspacing}{\spaceskip=\fontdimen2\font plus
\BIBentryALTinterwordstretchfactor\fontdimen3\font minus
  \fontdimen4\font\relax}
\providecommand{\BIBforeignlanguage}[2]{{%
\expandafter\ifx\csname l@#1\endcsname\relax
\typeout{** WARNING: IEEEtran.bst: No hyphenation pattern has been}%
\typeout{** loaded for the language `#1'. Using the pattern for}%
\typeout{** the default language instead.}%
\else
\language=\csname l@#1\endcsname
\fi
#2}}
\providecommand{\BIBdecl}{\relax}
\BIBdecl

\bibitem{sheridan1989telerobotics}
T.~B. Sheridan, ``Telerobotics,'' \emph{Automatica}, vol.~25, no.~4, pp.
  487--507, 1989.

\bibitem{tang2023csgp}
Z.~Tang \emph{et~al.}, ``Csgp: Closed-loop safe grasp planning via
  attention-based deep reinforcement learning from demonstrations,'' \emph{IEEE
  RA-L}, vol.~8, no.~6, pp. 3158--3165, 2023.

\bibitem{chi2023diffusion}
C.~Chi \emph{et~al.}, ``Diffusion policy: Visuomotor policy learning via action
  diffusion,'' \emph{IJRR}, vol.~44, no. 10-11, pp. 1684--1704, 2023.

\bibitem{kim2024openvla}
M.~J. Kim \emph{et~al.}, ``Openvla: An open-source vision-language-action
  model,'' \emph{CoRL}, 2025.

\bibitem{rouxel2025extremumflowmatching}
Q.~Rouxel \emph{et~al.}, ``Extremum flow matching for offline goal conditioned
  reinforcement learning,'' in \emph{IEEE-RAS Humanoids}, 2025.

\bibitem{schwarz2023robust}
M.~Schwarz \emph{et~al.}, ``Robust immersive telepresence and mobile
  telemanipulation: Nimbro wins ana avatar xprize finals,'' in \emph{IEEE-RAS
  Humanoids}, 2023.

\bibitem{zhao2023learning}
T.~Z. Zhao \emph{et~al.}, ``Learning fine-grained bimanual manipulation with
  low-cost hardware,'' \emph{RSS}, 2023.

\bibitem{ding2024bunny}
R.~Ding \emph{et~al.}, ``Bunny-visionpro: Real-time bimanual dexterous
  teleoperation for imitation learning,'' \emph{CoRR}, 2024.

\bibitem{liu2025avr}
Y.~Liu \emph{et~al.}, ``Avr: Active vision-driven robotic precision
  manipulation with viewpoint and focal length optimization,'' \emph{arXiv
  preprint arXiv:2503.01439}, 2025.

\bibitem{darvish2023teleoperation}
K.~Darvish \emph{et~al.}, ``Teleoperation of humanoid robots: A survey,''
  \emph{IEEE Transactions on Robotics}, vol.~39, no.~3, pp. 1706--1727, 2023.

\bibitem{rouxel2022multicontact}
Q.~Rouxel \emph{et~al.}, ``Multicontact motion retargeting using whole-body
  optimization of full kinematics and sequential force equilibrium,''
  \emph{IEEE/ASME TMECH}, vol.~27, no.~5, pp. 4188--4198, 2022.

\bibitem{mandlekar2018roboturk}
A.~Mandlekar \emph{et~al.}, ``Roboturk: A crowdsourcing platform for robotic
  skill learning through imitation,'' in \emph{CoRL}, 2018.

\bibitem{ozdamar2022shared}
I.~Ozdamar \emph{et~al.}, ``A shared autonomy reconfigurable control framework
  for telemanipulation of multi-arm systems,'' \emph{IEEE RA-L}, vol.~7, no.~4,
  pp. 9937--9944, 2022.

\bibitem{lin2024learning}
T.~Lin \emph{et~al.}, ``Learning visuotactile skills with two multifingered
  hands,'' \emph{IEEE ICRA}, 2024.

\bibitem{fung2005case}
W.-k. Fung \emph{et~al.}, ``A case study of 3d stereoscopic vs. 2d monoscopic
  tele-reality in real-time dexterous teleoperation,'' in \emph{IEEE/RSJ IROS},
  2005.

\bibitem{livatino2009stereo}
S.~Livatino \emph{et~al.}, ``Stereo viewing and virtual reality technologies in
  mobile robot teleguide,'' \emph{IEEE Trans. on Robotics}, vol.~25, no.~6, pp.
  1343--1355, 2009.

\bibitem{ishiguro2018high}
Y.~Ishiguro \emph{et~al.}, ``High speed whole body dynamic motion experiment
  with real time master-slave humanoid robot system,'' in \emph{IEEE ICRA},
  2018.

\bibitem{penco2019multimode}
L.~Penco \emph{et~al.}, ``A multimode teleoperation framework for humanoid
  loco-manipulation: An application for the icub robot,'' \emph{IEEE RAM},
  vol.~26, no.~4, pp. 73--82, 2019.

\bibitem{cheng2024open}
X.~Cheng \emph{et~al.}, ``Open-television: Teleoperation with immersive active
  visual feedback,'' \emph{CoRL}, 2024.

\bibitem{ryu1991functional}
J.~Ryu \emph{et~al.}, ``Functional ranges of motion of the wrist joint,''
  \emph{The Journal of hand surgery}, vol.~16, no.~3, pp. 409--419, 1991.

\bibitem{mehta2016integrating}
R.~K. Mehta, ``Integrating physical and cognitive ergonomics,'' \emph{IISE
  Trans. Occup. Ergon. Hum. Factors}, vol.~4, no. 2-3, pp. 83--87, 2016.

\bibitem{el2019survey}
F.~El~Jamiy \emph{et~al.}, ``Survey on depth perception in head mounted
  displays: distance estimation in virtual reality, augmented reality, and
  mixed reality,'' \emph{IET Image Processing}, vol.~13, no.~5, pp. 707--712,
  2019.

\bibitem{bressa2021s}
N.~Bressa \emph{et~al.}, ``What's the situation with situated visualization? a
  survey and perspectives on situatedness,'' \emph{IEEE TVCG}, vol.~28, no.~1,
  pp. 107--117, 2021.

\bibitem{martinez2019openpose}
G.~H. Mart{\i}nez, ``Openpose: Whole-body pose estimation,'' \emph{Ph.D.
  thesis}, 2019.

\bibitem{roetenberg2009xsens}
D.~Roetenberg \emph{et~al.}, ``Xsens mvn: Full 6dof human motion tracking using
  miniature inertial sensors,'' \emph{Xsens Motion Technologies BV, Tech. Rep},
  2009.

\bibitem{audonnet2024immertwin}
F.~P. Audonnet \emph{et~al.}, ``Immertwin: A mixed reality framework for
  enhanced robotic arm teleoperation,'' \emph{arXiv preprint arXiv:2409.08964},
  2024.

\bibitem{patil2024radiance}
V.~Patil \emph{et~al.}, ``Radiance fields for robotic teleoperation,'' in
  \emph{IEEE/RSJ IROS}, 2024.

\bibitem{laghi2018shared}
M.~Laghi \emph{et~al.}, ``Shared-autonomy control for intuitive bimanual
  tele-manipulation,'' in \emph{IEEE-RAS Humanoids}, 2018.

\bibitem{iyer2024open}
A.~Iyer \emph{et~al.}, ``Open teach: A versatile teleoperation system for
  robotic manipulation,'' \emph{CoRL}, 2025.

\bibitem{wen2023collaborative}
R.~Wen \emph{et~al.}, ``Collaborative bimanual manipulation using optimal
  motion adaptation and interaction control: Retargeting human commands to
  feasible robot control references,'' \emph{IEEE RAM}, vol.~31, no.~4, pp.
  68--80, 2023.

\bibitem{ajoudani2012tele}
A.~Ajoudani \emph{et~al.}, ``Tele-impedance: Teleoperation with impedance
  regulation using a body--machine interface,'' \emph{IJRR}, vol.~31, no.~13,
  pp. 1642--1656, 2012.

\bibitem{yang2024ace}
S.~Yang \emph{et~al.}, ``Ace: A cross-platform visual-exoskeletons system for
  low-cost dexterous teleoperation,'' \emph{CoRL}, 2025.

\bibitem{kanoun2011kinematic}
O.~Kanoun \emph{et~al.}, ``Kinematic control of redundant manipulators:
  Generalizing the task-priority framework to inequality task,'' \emph{IEEE
  T-RO}, 2011.

\bibitem{wen2022effects}
Z.~Wen \emph{et~al.}, ``Effects of view layout on situated analytics for
  multiple-view representations in immersive visualization,'' \emph{IEEE TVCG},
  vol.~29, no.~1, pp. 440--450, 2022.

\bibitem{white2009interaction}
S.~M. White, \emph{Interaction and presentation techniques for situated
  visualization}.\hskip 1em plus 0.5em minus 0.4em\relax Columbia University,
  2009.

\bibitem{chen2025unified}
Y.~Chen \emph{et~al.}, ``Unified model predictive interaction control
  integrating impedance matching and constraint optimization,'' in \emph{IEEE
  ICCA}, 2025.

\end{thebibliography}

\end{document}